\documentclass[11pt,a4paper]{article}
\usepackage{authblk}
\usepackage{stackengine}
\usepackage[hyperref]{emnlp2018}
\aclfinalcopy
\usepackage{times}
\usepackage{latexsym}

\usepackage{url}
\usepackage{xcolor}
\usepackage{upgreek}
\usepackage[shortlabels]{enumitem}
\usepackage{tabu}
\setlist[enumerate]{nosep}
\usepackage{floatrow}
\usepackage{multirow}
\usepackage{makecell}
\usepackage{amsmath}

\usepackage{booktabs}
\usepackage{graphicx}
\usepackage{amsfonts}
\usepackage{tabularx}
\usepackage{bm}

\def\aclpaperid{***}
\DeclareFloatFont{tiny}{\tiny}% "scriptsize" is defined by floatrow, "tiny" not
\usepackage{silence}
\usepackage{tabularx}
\usepackage{pifont}% http://ctan.org/pkg/pifont
\usepackage{subcaption}

\newlength\Origarrayrulewidth
\newcommand{\Cline}[1]{%
  \noalign{\global\setlength\Origarrayrulewidth{\arrayrulewidth}}%
  \noalign{\global\setlength\arrayrulewidth{1.1pt}}\cline{#1}%
  \noalign{\global\setlength\arrayrulewidth{\Origarrayrulewidth}}%
}

\usepackage{todonotes}
\usepackage{amsmath} % theoremstyls
\usepackage{wrapfig}
\DeclareMathOperator*{\bilstm}{\textit{BiLSTM}}
\DeclareMathOperator*{\mlp}{\textit{MLP}}
\renewcommand{\footnotesize}{\fontsize{8pt}{6pt}\selectfont}

\usepackage{graphicx}
\graphicspath{ {./images/} }
\usepackage[ruled]{algorithm2e}
\usepackage{amsthm}
\usepackage{amssymb}

\makeatletter
\newcommand{\vo}{\vec{o}\@ifnextchar{^}{\,}{}}
\makeatother

\title{KGPool: Dynamic Knowledge Graph Context Selection for Relation Extraction}

\author{Abhishek Nadgeri$^{1,2}$, Anson Bastos$^{1,3}$, Kuldeep Singh$^{1,5}$, Isaiah Onando Mulang'$^{1,7}$ \\
\textbf{Johannes Hoffart$^6$, Saeedeh Shekarpour$^4$, and Vijay Saraswat$^{6}$} \\
$^1$Zerotha Research, 
$^2$RWTH Aachen, 
$^3$IIT Hyderabad,
$^4$University of Dayton\\
$^5$Cerence GmbH, 
$^6$Goldman Sachs,
$^7$IBM Research, Africa\\
{\tt abhishek.nadgeri@rwth-aachen.de, cs20resch11002@iith.ac.in} \\
{\tt kuldeep.singh1@cerence.com, mulang.onando@ibm.com} \\
{\tt sshekarpour1@udayton.edu,\{johannes.hoffart,vijay.saraswat\}@gs.com}
  }

\date{}

\begin{document}
\setlength{\abovedisplayskip}{3pt}
\setlength{\belowdisplayskip}{3pt}

\maketitle

\begin{abstract}
We present a novel method for relation extraction (RE) from a single sentence, mapping the sentence and two given entities to a canonical fact in a knowledge graph (KG). Especially in this presumed sentential RE setting, the context of a single sentence is often sparse. This paper introduces the \emph{KGPool} method to address this sparsity, dynamically expanding the context with additional facts from the KG. It learns the representation of these facts (entity alias, entity descriptions, etc.) using neural methods, supplementing the sentential context. Unlike existing methods that statically use all expanded facts, \emph{KGPool} conditions this expansion on the sentence. We study the efficacy of \emph{KGPool} by evaluating it with different neural models and KGs (Wikidata and Freebase). Our experimental evaluation on standard datasets shows that by feeding the \emph{KGPool} representation into a Graph Neural Network, the overall method is significantly more accurate than state-of-the-art methods. 
\end{abstract}

%%%%%%%%%%%%%%%%%%%%%%%%%%%%%%%%%%%%%%%%%%%%%%%%%%%%%%%%%%%%%%%%%%%%
%%%%%%%%%%%%%%%%%%%%%%%%%%%%%%%%%%%%%%%%%%%%%%%%%%%%%%%%%%%%%%%%%%%%
%%%%%%%%%%%%%%%%%%%%%%%%%%%%%%%%%%%%%%%%%%%%%%%%%%%%%%%%%%%%%%%%%%%%
%%%%%%%%%%%%%%%%%%%%%%%%%%%%%%%%%%%%%%%%%%%%%%%%%%%%%%%%%%%%%%%%%%%%
%%%%%%%%%%%%%%%%%%%%%%%%%%%%%%%%%%%%%%%%%%%%%%%%%%%%%%%%%%%%%%%%%%%%
%%%%%%%%%%%%%%%%%%%%%%%%%%%%%%%%%%%%%%%%%%%%%%%%%%%%%%%%%%%%%%%%%%%%
%%%%%%%%%%%%%%%%%%%%%%%%%%%%%%%%%%%%%%%%%%%%%%%%%%%%%%%%%%%%%%%%%%%%
%%%%%%%%%%%%%%%%%%%%%%%%%%%%%%%%%%%%%%%%%%%%%%%%%%%%%%%%%%%%%%%%%%%%
%%%%%%%%%%%%%%%%%%%%%%%%%%%%%%%%%%%%%%%%%%%%%%%%%%%%%%%%%%%%%%%%%%%%
%%%%%%%%%%%%%%%%%%%%%%%%%%%%%%%%%%%%%%%%%%%%%%%%%%%%%%%%%%%%%%%%%%%%

\section{Introduction} \label{sec:introduction}
Knowledge graphs (KGs) are the foundation for many downstream applications and are growing ever larger. However, due to the sheer volume of knowledge and the world's dynamic nature where new entities emerge and unknown facts about them are learned, KGs need to be continuously updated.
Distantly supervised Relation Extraction (RE) is an important KG completion task aiming at finding a semantic relationship between two entities annotated on the unstructured text with respect to an underlying knowledge graph \cite{DBLP:conf/naacl/YeL19}.
In the literature, researchers mainly studied two variants in the RE: 1) multi-instance RE and 2) sentential RE. The multi-instance RE assumes that in a given bag of sentences, if two entities participate in a relation, there exists at least one sentence with these
two entities, which may contain the target relation \cite{DBLP:conf/pkdd/RiedelYM10,DBLP:conf/emnlp/VashishthJPBT18}.
In this setting, researchers aim to incorporate contextual signals from the previous occurrences of an entity pair into the neural models to support relation extraction \cite{DBLP:conf/naacl/YeL19,DBLP:conf/naacl/XuB19,DBLP:conf/aaai/WuFZ19}.
%\todo[inline]{SS: instead of previous better to say contexual}
%\cite{DBLP:conf/aaai/WuFZ19}. In contrast, sentential RE limits the available context to the input sentence (ignoring previous occurrences of entity pair in the bag of sentences) while predicting the KG relation \cite{DBLP:conf/emnlp/SorokinG17}.
In contrast, sentential RE restricts the scope of document context only to the input sentence (disregards other occurrences of entity pairs) while predicting the KG relation \cite{DBLP:conf/emnlp/SorokinG17}. Hence, sentential RE makes the multi-instance setting more difficult by limiting the available context.\\
\begin{figure}[t]
\includegraphics[scale = 0.57]{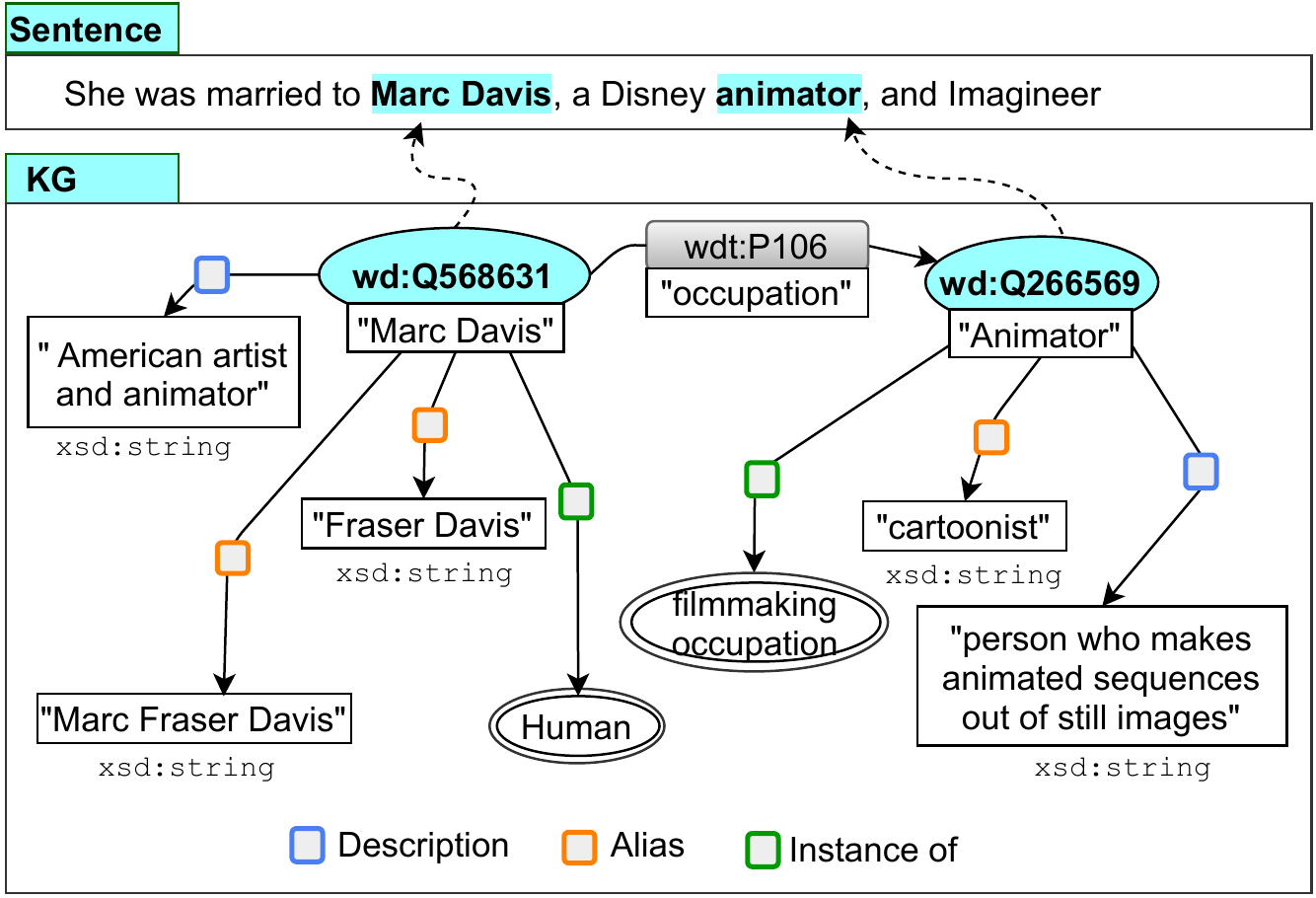}
\centering
\caption{Illustration of Knowledge Graph Context associated with the annotated entities in a sentence. Here, entity aliases do not play any role in the understanding of the sentence for finding the KG relation. \footnote{sentence taken from \cite{DBLP:conf/emnlp/SorokinG17}}}
\label{fig:motivation}
\end{figure}
%\todo[inline]{SS: Figure 1: what is aim of this example? do you want to illustrate that the relation of "occupation" has be inferred? if yes, it is not clear how and which part of context helped for this inference?}
%Recent approaches for RE not only use KGs as the repository of target relation but also use it as a sourceof knowledge to improve the result of the RE\cite{DBLP:conf/emnlp/VashishthJPBT18,bastos2020recon}.
Recent approaches for RE not only base KGs for relation inventory but also consider it for extending contextual knowledge for further improvement of RE task \cite{DBLP:conf/emnlp/VashishthJPBT18,bastos2020recon}.
%A few multi-instance RE methods use entity attributes (properties) such as descriptions, aliases, and types along with entity pair occurrences from previous sentences to improve the overall extraction quality \cite{DBLP:conf/aaai/Ji0H017,DBLP:conf/emnlp/VashishthJPBT18}.
A few multi-instance RE methods rely on entity attributes (properties) such as descriptions, aliases, and types (as additional context) along with entity pair occurrences from previous sentences to improve the overall extraction quality \cite{DBLP:conf/aaai/Ji0H017,DBLP:conf/emnlp/VashishthJPBT18}. For the sentential RE,
the RECON approach \cite{bastos2020recon} aims to effectively encode KG context derived from the entity attributes and entity neighborhood triples. RECON employs a Graph Neural Network (GNN) as a context aggregator for combining sentential context (annotated entities and sentence) and structured KG representation. Although the additional KG context has a positive effect on the overall relation extraction in multi-instance and sentential RE settings, not all KG context forms are necessary for every input sentence. 
%\todo[inline]{it concludes that the not all KG context are necessary for every input sentence}
Consider Figure \ref{fig:motivation}, where the task is to infer a semantic relation ‘occupation' between two entities \textit{wd:Q568631 (Marc Davis)}\footnote{wd: binds to \url{https://www.wikidata.org/wiki/}} and \textit{wd:Q266569 (animator)}. Wikidata \cite{DBLP:conf/www/Vrandecic12} KG provides semantic information such as description, instance-of, and aliases about entities.
%Wikidata \cite{DBLP:conf/www/Vrandecic12} provides semantic information such as about entities in the form of their attributes such as description, alias, etc. 
Here, the entity alias (Marc Fraser Davis, Fraser Davis for \textit{wd:Q568631}; and cartoonist for \textit{wd:Q266569}) has no impact on understanding the sentence because the entities are explicitly mentioned in the sentence. Furthermore, there is empirical evidence in the literature that for several sentences, statically adding all KG context offered minimal or negative impact \cite{bastos2020recon}.
%\todo[inline]{SS: adding all KG context might cause minimal or negative impact}
%Hence, it is an open research question of how an approach should dynamically select a specific form of the KG context? Does it positively impact the overall performance? 
Hence, there are open research questions as to how an RE approach can dynamically utilize the sufficient context from KG and whether or not the selected KG context positively impacts the overall performance?

This paper studies these concerning questions
proposing the \textit{KGPool} approach. \textit{KGPool} utilizes a self-attention mechanism in a Graph Convolution Network (GCN) \cite{kipf2016semi} for selecting a sub-graph from the KG to extend the sentential context. The concept of dynamically mapping the structural representation of a KG to a latent representation of a sentence has not been widely studied in prior literature. In RE, \textit{KGPool} is the initial attempt. The existing approaches \cite{bastos2020recon,DBLP:conf/naacl/XuB19,DBLP:conf/aaai/WuFZ19,DBLP:conf/emnlp/VashishthJPBT18} feed all the available context (either derived from a bag of sentences or a KG or both) into a neural model and relied on the model to figure out the consequences, resulting in limited performance in many cases \cite{bastos2020recon}. 
Conversely, we study the efficacy of \textit{KGPool} in dynamically choosing KG context for the sentential RE task using two standard community datasets (NYT Freebase \cite{DBLP:conf/pkdd/RiedelYM10}, and Wikidata \cite{DBLP:conf/emnlp/SorokinG17}). Our work makes the following key contributions:
\begin{itemize}
\itemsep-.4em 
  \item The \textit{KGPool} approach dynamically selects structural knowledge and transform it into a representation suitable to supplement the latent representation of sentential context learned using a neural model. We deduce that \textit{KGPool} is the first approach that works independently of the underlying context aggregators used in the literature (Graph Neural Network \cite{DBLP:conf/acl/ZhuLLFCS19} or LSTM-based \cite{DBLP:conf/emnlp/SorokinG17}).
  \item We are the first to map the task of KG Context Selection to a Graph Pooling Problem. Therefore, our proposed approach legitimizes the application of graph pooling algorithms for choosing the relevant context. 
  \item \textit{KGPool}, paired with a GNN as context aggregator, significantly outperforms the existing baselines on both datasets, in one experiment increasing the precision by 12 points over to baseline (P@30 on NYT Freebase). Furthermore, our empirical results (cf., Table \ref{tab:results2}) conclude that an LSTM model paired with \textit{KGPool} is able to notably outperform a GNN-based approach \cite{DBLP:conf/acl/ZhuLLFCS19} and nearly all multi-instance baselines \cite{DBLP:conf/naacl/YeL19,DBLP:conf/aaai/WuFZ19,DBLP:conf/emnlp/VashishthJPBT18} published in the recent years. 
\end{itemize}
%Our exhaustive evaluations conclude that \textit{KGPool} coupled with a graph neural network as context aggregator significantly outperforms the existing baselines on both datasets. In some cases, the jump is five absolute points compared to baseline (P@10 at NYT Freebase).
This paper is structured as follows: Section \ref{sec:related} reviews the related work. Section \ref{sec:problem} formalizes the problem and the proposed approach is described in Section \ref{sec:approach}. Section \ref{sec:experiment} describes the experimental setup. The results are in Section \ref{sec:results}. We conclude in Section \ref{sec:conclusion}.

%%%%%%%%%%%%%%%%%%%%%%%%%%%%%%%%%%%%%%%%%%%%%%%%%%%%%%%%%%%%%%%%%%%%%%%%%%%%%%%%%
%%%%%%%%%%%%%%%%%%%%%%%%%%%%%%%%%%%%%%%%%%%%%%%%%%%%%%%%%%%%%%%%%%%%%%%%%%%%%%%%%
%%%%%%%%%%%%%%%%%%%%%%%%%%%%%%%%%%%%%%%%%%%%%%%%%%%%%%%%%%%%%%%%%%%%%%%%%%%%%%%%%
%%%%%%%%%%%%%%%%%%%%%%%%%%%%%%%%%%%%%%%%%%%%%%%%%%%%%%%%%%%%%%%%%%%%%%%%%%%%%%%%%
%%%%%%%%%%%%%%%%%%%%%%%%%%%%%%%%%%%%%%%%%%%%%%%%%%%%%%%%%%%%%%%%%%%%%%%%%%%%%%%%%
%%%%%%%%%%%%%%%%%%%%%%%%%%%%%%%%%%%%%%%%%%%%%%%%%%%%%%%%%%%%%%%%%%%%%%%%%%%%%%%%%
%%%%%%%%%%%%%%%%%%%%%%%%%%%%%%%%%%%%%%%%%%%%%%%%%%%%%%%%%%%%%%%%%%%%%%%%%%%%%%%%%
%%%%%%%%%%%%%%%%%%%%%%%%%%%%%%%%%%%%%%%%%%%%%%%%%%%%%%%%%%%%%%%%%%%%%%%%%%%%%%%%%
\section{Related Work} \label{sec:related}
%in the paradigm of distant supervision (weakly supervised) RE, multi-instance setting assumes that if two entities engage in a relation, at least one sentence that mentions those two entities might express that relation. 
\textbf{Multi-instance RE:} a few multi-instance RE approaches use convolution neural network \cite{dos2015classifying}, attention CNN \cite{wang2016relation} and attention-based recurrent neural models for relation extraction \cite{zhou2016attention}. Other approaches such as \cite{DBLP:conf/aaai/Ji0H017,DBLP:conf/emnlp/VashishthJPBT18} incorporate entity descriptions, entity and relation aliases from KG to supplement context from the previous sentences. Work in \cite{DBLP:conf/emnlp/VashishthJPBT18} employs a graph convolution network to encode entity and relation aliases derived from Wikidata. HRERE \cite{DBLP:conf/naacl/XuB19} proposes an approach for jointly learning sentence and KG representation using cross-entropy loss function. To effectively capture the available entity context in the documents,  \citet{DBLP:conf/naacl/YeL19} suggest an approach incorporating intra-bag and inter-bag attentions.
For a detailed survey, we point readers to \cite{smirnova2018relation}.\\
\textbf{Sentential RE:} researchers \cite{DBLP:conf/emnlp/SorokinG17} utilized additional relations present in the sentence to assist the process of extracting the target relation using an LSTM-based model. GP-GNN \cite{DBLP:conf/acl/ZhuLLFCS19} generates parameters of GNN based on the input sentence, which enables GNNs to process-relational reasoning
on unstructured text inputs. RECON \cite{bastos2020recon} is an approach that uses the entity attributes (aliases, labels, descriptions, instance-of) and KG triples to signal an underlying GNN model for sentential RE. 
%The paper presents a model that obtains strong results in the RE task against other sentential models as well as against multi-instance models.
Authors conclude that the multi-instance requirement can be relaxed provided a good representation of KG context to enrich the sentential RE model. However, RECON and multi-instance approaches \cite{DBLP:conf/naacl/XuB19,DBLP:conf/emnlp/VashishthJPBT18} utilize statically derived context from the KG, i.e., KG context does not vary depending the sentence.

%Coke-BERT is a recent approach that dynamically select the KG context using a GNN model to enhance the performance of the underlying transformer model
\textbf{Graph Pooling and Dynamic Context Selection}: researchers proposed several models for the graph classification aka. graph pooling task \cite{cangea2018towards,ying2018hierarchical,gao2019graph}. These models employ various approaches such as graph topology-based \cite{rhee2018hybrid}, and by learning the hierarchical graph-structure \cite{ying2018hierarchical}. Another graph pooling model relies on node features and topological information using self-attention \cite{lee2019self} in which a specific number of nodes are always eliminated. In \textit{KGPool}, the elimination of nodes depends on a context coefficient and node importance (Section \ref{sec:approach}). For context selection, a recent work focuses on dynamically selecting the KG context to optimize a Pre-Trained Language Model (PLM) for entity typing and relation classification \cite{su2020contextual}. \textit{KGPool} has the following fundamental differences compared to \cite{su2020contextual}: \textit{KGPool} inspires its self-attention mechanism from \cite{lee2019self,vaswani2017attention} to learn a representation of the KG context. Hence, \textit{KGPool} works agnostic of the underlying model used for the context aggregation (unlike \citet{su2020contextual}, which is tightly coupled with PLM). Approaches such as \cite{liu2017context,DBLP:conf/emnlp/Zhang0M18,DBLP:conf/emnlp/KangZZZ20} also perform dynamic context selection for respective tasks.  However, these approaches are not focused on knowledge graph context selection.
%%%%%%%%%%%%%%%%%%%%%%%%%%%%%%%%%%%%%%%%%%%%%%%%%%%%%%%%%%%%%%%%%%%%%%%%%%%%%%%%%
%%%%%%%%%%%%%%%%%%%%%%%%%%%%%%%%%%%%%%%%%%%%%%%%%%%%%%%%%%%%%%%%%%%%%%%%%%%%%%%%%
%%%%%%%%%%%%%%%%%%%%%%%%%%%%%%%%%%%%%%%%%%%%%%%%%%%%%%%%%%%%%%%%%%%%%%%%%%%%%%%%%
%%%%%%%%%%%%%%%%%%%%%%%%%%%%%%%%%%%%%%%%%%%%%%%%%%%%%%%%%%%%%%%%%%%%%%%%%%%%%%%%%
%%%%%%%%%%%%%%%%%%%%%%%%%%%%%%%%%%%%%%%%%%%%%%%%%%%%%%%%%%%%%%%%%%%%%%%%%%%%%%%%%
%%%%%%%%%%%%%%%%%%%%%%%%%%%%%%%%%%%%%%%%%%%%%%%%%%%%%%%%%%%%%%%%%%%%%%%%%%%%%%%%%
%%%%%%%%%%%%%%%%%%%%%%%%%%%%%%%%%%%%%%%%%%%%%%%%%%%%%%%%%%%%%%%%%%%%%%%%%%%%%%%%%
%%%%%%%%%%%%%%%%%%%%%%%%%%%%%%%%%%%%%%%%%%%%%%%%%%%%%%%%%%%%%%%%%%%%%%%%%%%%%%%%%
\section{Problem Statement}\label{sec:problem}
\begin{figure*}
	\centering
	\includegraphics[width=\textwidth]{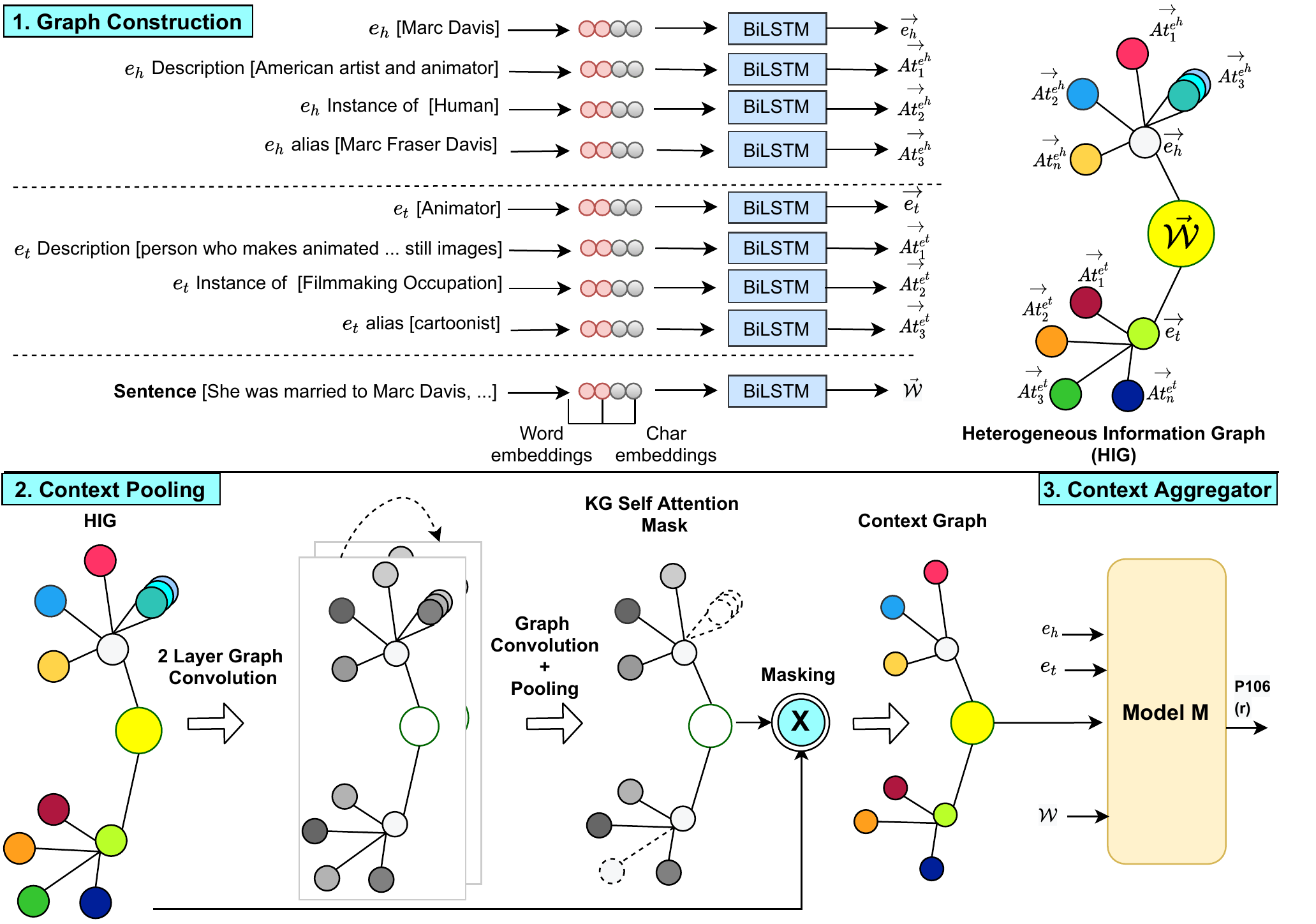}
	\caption{\textit{KGPool} approach has three components to supplement sentential context with necessary KG context.}
%\caption{\textit{KGPool} approach has three components to supplement sentential context with necessary KG context. The Graph Construction step forms a Heterogeneous Information Graph ($HIG$) for input representation. In Context Pooling, we utilize graph convolution coupled with a pooling layer to generate Context Graph ($CG$), a sub-graph of $HIG$ by removing less important entity attribute nodes. The last step aggregates the chosen KG context ($CG$) with the latent representation of sentential context into a Context Aggregator to predict the final relation.
	\label{fig:approach}
	    \vspace{-2mm}
	\end{figure*}

%Formalization of problem,  
%definitions \\
%\subsection{Problem Statement}
%%%%%%%%%%%% %%%%%%%%%%%%%%%%%%%%%%%%%%%%%%%%%%%
We define the KG as a tuple given by $KG = (\mathcal{E},\mathcal{R},\mathcal{T}^+)$ where $\mathcal{E}$ denotes the set of all vertices in the graph representing entities, $\mathcal{R}$ is the set of edges representing relations, and $\mathcal{T}^+ \subseteq \mathcal{E} \times \mathcal{R} \times \mathcal{E} $ is a set of all KG triples. The \textit{RE Task} predicts the target relation $r^c \in \mathcal{R}$ between a given pair of entities $\langle e_i,e_j\rangle$ from the sentence $\mathcal{W} = ( w_1,w_2,...,w_l )$.
%\todo[inline]{The definition of the sentence is missing, and it should be before the definition of RE task, maybe you can define a function which maps input sentence to a triple}
If no relation is inferred, it returns 'NA' label. We aim for the \underline{sentential RE} task which put a constraint that the sentence within which a given pair of entities occurs is the only visible sentence from the bag of sentences. 
%\todo[inline]{SS: this sentence is unclear and even I think unnecessary: We aim for the \underline{sentential RE} task which put a constraint that the sentence within which a given pair of entities occurs is the only visible sentence from the bag of sentences. }
%The sentential RE is a single label RE task.
%\todo[inline]{SS: what do you mean by "single label"? }
%In sentential RE, no context is considered from previous occurrences of the entities present in the bag.
%We aim for the \underline{sentential RE} task which put a constraint that the sentence within which a given pair of entities occurs is the only visible sentence from the bag of sentences. In sentential RE, no context is considered from previous occurrences of the entities present in the bag.
%All other sentences in the bag are not considered while predicting the correct relation $r^c$.
We view RE as a classification task similar to ~\cite{DBLP:conf/emnlp/SorokinG17}. 
%For the same \todo[inline]{instead of For the same, similarly}, we define a sentence which is a sequence of words as $\mathcal{W} = ( w_1,w_2,...,w_l )$. 
In a KG triple $\uptau = (e_h,r,e_t) \in \mathcal{T}^+$, the relation $r \in \mathcal{R}$, $e_h$ is the head entity (relation origin) while $e_t$ is the tail entity. For each entity, associated semantic properties such as entity label, description, instance-of, and aliases are known as entity attribute ($At^e$) (cf., graph construction step of Figure \ref{fig:approach}).
We aim to model KG contextual information to improve the classification. This is achieved by learning the effective representations of the sets $At^e$, $e_h,e_t$, and $\mathcal{W}$ (cf. section \ref{sec:approach}).

% relevance
%%%%%%%%%%%%%%%%%%%%%%%%%%%%%%%%%%%%%%%%%%%%%%%%%%%%%%%%%%%%%%%%%%%%%%%%%%%%%%%%%
%%%%%%%%%%%%%%%%%%%%%%%%%%%%%%%%%%%%%%%%%%%%%%%%%%%%%%%%%%%%%%%%%%%%%%%%%%%%%%%%%
%%%%%%%%%%%%%%%%%%%%%%%%%%%%%%%%%%%%%%%%%%%%%%%%%%%%%%%%%%%%%%%%%%%%%%%%%%%%%%%%%
%%%%%%%%%%%%%%%%%%%%%%%%%%%%%%%%%%%%%%%%%%%%%%%%%%%%%%%%%%%%%%%%%%%%%%%%%%%%%%%%%
%%%%%%%%%%%%%%%%%%%%%%%%%%%%%%%%%%%%%%%%%%%%%%%%%%%%%%%%%%%%%%%%%%%%%%%%%%%%%%%%%
%%%%%%%%%%%%%%%%%%%%%%%%%%%%%%%%%%%%%%%%%%%%%%%%%%%%%%%%%%%%%%%%%%%%%%%%%%%%%%%%%
%%%%%%%%%%%%%%%%%%%%%%%%%%%%%%%%%%%%%%%%%%%%%%%%%%%%%%%%%%%%%%%%%%%%%%%%%%%%%%%%%
%%%%%%%%%%%%%%%%%%%%%%%%%%%%%%%%%%%%%%%%%%%%%%%%%%%%%%%%%%%%%%%%%%%%%%%%%%%%%%%%%

\section{KGPool Approach} \label{sec:approach}
\textit{KGPool} consists of three components illustrated in Figure \ref{fig:approach}: 1) \textit{Graph Construction} aggregates the sentence, entities and its attributes as a Heterogeneous Information Graph ($HIG$) for input representation.
2) \textit{Context Pooling} step utilizes a self-attention mechanism in a graph convolution to calculate attention scores for entity attributes using node features and graph topology. The pooling process allows \textit{KGPool} to construct a Context Graph ($CG$), which is a contextualized representation of $HIG$ with lesser number of nodes.
3) \textit{Context Aggregator} takes as input the sentence, entities, contextual representations of $HIG$, and classifies the target relation between the entities.
%We now detail the approach.
We detail the approach in the following.
%The KGPool approach is divided into three parts. The first step is to create the information graph (sec. \ref{sec:graph_construction}). It is followed by creating the context graph through KG Context Pooling  (sec. \ref{sec:pooling}). The pooled graph is then fed into the KG aggrgator (sec. \ref{sec:aggregator}).
\subsection{Graph Construction} \label{sec:graph_construction}
As first step, we extract entity attributes ($At^e$) from public dumps of Freebase \cite{DBLP:conf/aaai/BollackerCT07} and Wikidata \cite{DBLP:conf/www/Vrandecic12} KGs depending on the dataset. For the KG context, we rely on the most commonly available properties of the entities as suggested by \cite{bastos2020recon}: aliases, description, instance-of, and label. An example of various entity attributes is given in Figure \ref{fig:approach} at the Graph Construction step. 
%Then, the sentence $\mathcal{W}$ is converted into its corresponding representation using Bi-LSTM
Then, the sentence $\mathcal{W}$ is transformed to another representation using Bi-LSTM \cite{schuster1997bidirectional} by concatenating its word and character embeddings: 
\begin{equation} \label{eq_1}
\vec{\mathcal{W}} = \bilstm (\mathcal{W})
\end{equation}
Similar representation is created for each entity $e_i$ where $i=(h,t)$:
\begin{equation} \label{eq_2}
\vec{e_i} = \bilstm (e_i)
\end{equation}
For entity $e_i$, its KG contexts (entity attributes) $At_j^{e^i}$ (where $j=[0...N]$) are independently converted into associated embedding representations:
\begin{equation} \label{eq_3}
\vec{At_j}^{e^i} = \bilstm (At_j^{e^i})
\end{equation}
For a knowledge representation of the KG context concerning the sentential context (sentence and annotated entities), we introduce the special graph $HIG=(A,F)$, a Heterogeneous Information Graph, represented in the adjacency matrix $A \in\{0,1\}^{n \times n}$, where $n$ is the maximum number of neighboring nodes for an entity $e_i$.
%\todo[inline]{SS: For representing context from KG, we introduce an special graph named Heterogeneous Information Graph and denoted by  $HIG=(A,F)$. This graph is represented in the adjacency matrix $A \in\{0,1\}^{n \times n}$, where $n$ is the maximum number of neighboring nodes for an entity $e_i$.}
Here, $F \in \mathbb{R}^{n \times f}$ is the node feature matrix 
%\todo[inline]{SS: node means what? does it refer to entity?if yes, better to use entity}
assuming each node has $f$ features learned from the Bi-LSTM in the equations \ref{eq_1}, \ref{eq_2}, and \ref{eq_3}. In these equations, BERT \cite{devlin2019bert}, or any other recent Transformer-based model can be used. Due to hardware limitations, we are bound to Bi-LSTM using Glove embeddings \cite{pennington2014glove}.
%%%%%%%%%%%%%%%%%%%%%%%%%%%%%%%%%%%%%%%%%%%%%%
%%%%%%%%%%%%
%%%%%%%%%%%%%%%%%%%%%
%%%%%%%%%%%%%%%%%%%%%%%%%%%%%%%%
%%%%%%%%%%%%%%%%%%%%%%%%
%%%%%%%%%%%%%%%%%%%%%%%%%%%%%%%%%
%%%%%%%%%%%%%
%%%%%%%%%%%%%%%%
\subsection{Context Pooling} \label{sec:context_pooling}
%Context pooling has three blocks each of which consists of a Graph Convolutional Layer (GCN) 
%\todo[inline]{GCN is not abbreviation of Graph Convolutional Layer, I fixed it in the revision }
Context pooling is built upon three layers of Graph Convolutional Networks (GCN)
and a readout layer associated with each of them. Moreover, the last layer of GCN is coupled with a pooling layer (cf., ablation studies for architectural design choice experiments).
%%%%%%%%%%%%%%%%%%%%%%%%
%%%%%%%%%%%%%%%%%%%%%%%%%%%%%%%%%
%%%%%%%%%%%%%
%%%%%%%%%%%%%%%%
%%%%%%%%%%%%%%%%%%%%%%%%
%%%%%%%%%%%%%%%%%%%%%%%%%%%%%%%%%
%%%%%%%%%%%%%
%%%%%%%%%%%%%%%%
\subsubsection{Graph Convolution}\label{sec:graph_convolution}
 %Considering \textit{KGPool} only chooses the KG context necessary for a given sentential context, 
 Since \textit{KGPool} is expected to select the sufficient context, the Context Graph $CG$ is a reduction of $HIG$ using the mapping $\Psi : HIG \longrightarrow CG$.  
The challenge here is the no natural notion of spatial locality, i.e., it is not viable to pool together all context nodes in an “m × m” patch on $HIG$ because the complex topological structure of graphs prevents any straightforward,
deterministic definition of a “patch”. Furthermore,
% \todo[inline]{SS: this phrase is not clear at all "The challenge here is the no natural notion of spatial locality"} i.e., it is not viable to pool together all context nodes in an “m × m” patch on $HIG$ because the graph \todo[inline]{SS: why not viable?? justification?}
entities nodes have a varying number of neighboring nodes, making the graph pooling challenging (similar to other graph classification problems \cite{ying2018hierarchical}).
% \todo[inline]{Since entities vary in the number of attributes, thus having a fixed schema for graph pooling is not possible.}
% \todo[inline]{SS: sometime you mention entity sometimes node, better to bind to one of them, or also explicitly mention somewhere}
In $HIG$, entity nodes do not contain information of their neighbors. Hence, we aim to enrich each entity node with the adjacent node's contextual information. 
%In $HIG$, nodes do not contain the information from their neighbors. 
%\todo[inline]{SS: In $HIG$, nodes do not contain information of their neighbors.}
%Hence, our idea is to enrich each node with the adjacent node's contextual information in the graph. 
%\todo[inline]{Hence, we aim to enrich each node with the adjacent node's contextual information. }
Therefore, we employ a GNN variant to utilize its message-passing architecture to learn node embeddings from a message propagation function. 
The message propagation function depends on the adjacency matrix $A$, trainable parameters $\theta$, and the node embeddings $F$ \cite{ying2018hierarchical}. We rely on a GCN model by \citet{kipf2016semi}. The GCN layer is defined as:  
\begin{equation} \label{eq_4}
F^{(k)} = \operatorname{ReLU}\left(\tilde{D}^{-\frac{1}{2}} \tilde{A} \tilde{D}^{-\frac{1}{2}} F^{(k-1)} \theta^{(k-1)}\right)
\end{equation} 
where $\tilde{A} = A + I$, $\tilde{D} = \sum_{j} \tilde{A_{i,j}}$ and $\theta^{(k)}$ is the trainable matrix. The GCN module might run 
%\todo[inline]{The GCN module run or (might run) ---}
$k$ iteration and normally is in the range of two to six \cite{ying2018hierarchical}. 
%We use the readout layer that aggregates node features to make a fixed size representation similar to \cite{xu2018representation,cangea2018towards}. We perform this summarizing after each block of the network (equation \ref{eq_4}), and aggregate all intermediate representations together by taking their sum. We define readout layer $R$ as:
A few graph representation learning approaches propose to use readout layer that aggregates node features to learn a fixed size representation \cite{xu2018representation,cangea2018towards}. We perform this summarizing after each block of the network (Equation \ref{eq_4}), and aggregate all of the intermediate representation together by taking their sum. We define readout layer $R$ as:
\begin{equation} \label{eq_4.1}
R^{(k)}=\frac{1}{N} \sum_{i=1}^{N} F^{(k)}_{i} \| \max _{i=1}^{N} F^{(k)}_{i}
\end{equation} 
where $N$ is the number of nodes in the graph and $F$ is the node feature embedding. 
%%%%%%%%%%%%%%%%%%%%%%%%%%%%%%%%%%%%%%%%%%%%%%%%%%%%%%%%%%%%%%%%%%%%%%%%%%%%5
%%%%%%%%%%%%%%%%%%%%%%%%%%%%%%%%
%%%%%%%%%%%%%%%%%%%%%%%%%%%%%%%%%%%%%%%%%%%%%%%%%%%%%%%%
%%%%%%%%%%%%%%%%%%%%%%%%%%%%%%%%%%%%%%%%%%%%%%%%%%%%%%%%%%%%%%%%%%%%%%%%
%%%%%%%%%%%%%%%%%%%%%%%%%%%%%%%%%%%%%%
\subsubsection{KG Self-Attention Mask} \label{sec:self_attention}
%Attention mechanism focus more on important features and less on unimportantfeatures making it a widely used concept in the literature \cite{DBLP:conf/aaai/Han0S18,DBLP:conf/naacl/YeL19}. 
%In case of self-attention, input features are also a criteria for the attention itself \cite{vaswani2017attention}. 
Until Equation \ref{eq_4.1}, \textit{KGPool} focuses on learning node features. Next, \textit{KGPool} learns the importance of each entity attribute node using self-attention. Please note, in $HIG$, pooling happens only for entity attribute nodes ($\vec{At_j}^{e^i}$ from Equation \ref{eq_3}). The sentence $\vec{\mathcal{W}}$ and entities $\vec{e_h},\vec{e_t}$ remain intact. Hence, each entity representation $\vec{e_h}$ and $\vec{e_t}$ is enriched by the useful attribute context (KG context). The entity attribute nodes which do not provide relevant context are excluded from the graph.
%The entity attribute nodes which are not required for the given input sentence gets removed from the graph. 
%\todo[inline]{The entity attribute nodes which do not provide relevant context are excluded from the graph.}
%\todo[inline]{question: what is the rational behind this exclusion decision? how such nodes are recognized}
To choose the relevant entity attribute nodes, we use a self-attention score $Z$ \cite{lee2019self} calculated as follows:
\begin{equation} \label{eq_5}
Z=\tanh\left(\tilde{D}^{-\frac{1}{2}} \tilde{A} \tilde{D}^{-\frac{1}{2}} F^{(k)}  \Theta_{a t t}\right)
\end{equation} 

where $\Theta_{a t t} \in \mathbb{R}^{F \times 1}$ is the only parameter of the pooling layer.
For ranking, we take the attention score and pass it through a softmax layer where  $Z_{score}$ is the normalized self attention score.
%\begin{equation} \label{eq_6}
\begin{equation} \label{eq_6}
Z_{score} = \exp{(Z_{i})}/ \sum_{i}\exp{(Z_{i})}
\end{equation} 
After Equation \ref{eq_6}, we compute the scores for each entity attribute node. Next, we propose a \underline{node selection method} in which nodes are selected on the basis of \textbf{Context Coefficient} $\alpha$ which is a hyper parameter. The top nodes are selected as:
\begin{equation} \label{eq_7}
idx = \max{(Z_{score})} - \alpha * \sigma{(Z_{score})} 
\end{equation} 
where $\sigma{(Z_{score})}$ is the standard deviation of $Z_{score}$, $idx$ represents the node selection result, and $Z_{mask}$ is the corresponding attention mask. Equation \ref{eq_7} acts as a soft constraint in selecting the context nodes for each $HIG$ which depends on the value of $\alpha$. Learning $\alpha$ during training may cause over-fitting. Hence, we decided to consider $\alpha$ as a trade-off parameter similar to $\lambda$ in regularization \cite{buhlmann2011statistics}.
%we decided to let $\alpha$ as a trade-off for the developer, same as we normally do for $\lambda$ in regularization \cite{buhlmann2011statistics}. 
%\todo[inline]{Hence, we decided to consider $\alpha$ as a trade-off parameter similar to $\lambda$ in regularization \cite{buhlmann2011statistics}.}
Next, the Context Graph ($CG$) is formed by pooling out the less essential entity attribute nodes as: 
\begin{equation} \label{eq_8}
F^{\prime}=F^{(k)}_{\mathrm{idx},:}, F_{o u t}=F^{\prime} \odot Z_{m a s k}, A_{o u t}=A_{\mathrm{idx}, \mathrm{idx}}
\end{equation}
In addition to the dynamically selected nodes, we also inherit the intermediate node and graph representations of $k-1$ layers similar to ResNET \cite{he2016deep}. The intermediate representations ($k-1$ ) and the $CG$ ($k^{th}$ layer) is given as follows: 
\begin{equation} \label{eq_9}
\begin{aligned}
\vec{e_h}^{\prime} = F^{(1)}_{e_h} \oplus F^{(2)}_{e_h} \oplus .... F^{(k)}_{e_h}\\
\vec{e_t}^{\prime} = F^{(1)}_{e_t} \oplus F^{(2)}_{e_t} \oplus .... F^{(k)}_{e_t}\\
\vec{\mathcal{W}}^{\prime} = F^{(1)}_{\mathcal{W}} \oplus F^{(2)}_{\mathcal{W}} \oplus ....  F^{(k)}_{\mathcal{W}}\\
\vec{R}^{\prime} = R^{(1)} \oplus R^{(2)} \oplus ....  R^{(k)}
\end{aligned}
\end{equation}
where in the $i^{th}$ layer: $F^{(i)}_{e_l}$ is the node embedding of $e_l$, $l = (h,t)$, $F^{(i)}_{\mathcal{W}}$ is the node embedding of sentence $\mathcal{W}$, and $R^{(i)}$ is the readout. In the $k^{th}$ layer, $F^{(k)}$ is the  $F_{o u t}$ from Equation \ref{eq_8}. The $\oplus$ is the concatenation among the vectors. 
%The representations of eq. \ref{eq_9} are then fed into context aggregator.
%%%%%%%%%%%%%%%%%%%%%%%%%%%
%%%%%%%%%%%%%%%%%%%%%%%%%%%
%%%%%%%%%%%%%%%%%%%%%%%%%%%%%%%
%%%%%%%%%%%
\subsection{Context Aggregator}  \label{sec:aggregator}
Finally, \textit{KGPool} combines the latent representation (sentential context) with the structured representation learned in Equation \ref{eq_9}. As such, we employ a model $M$ which learns latent
relation vector $\vec{r^{\prime}}$. In the state-of-the-art approaches that use KG context, the representation of $\vec{r^{\prime}}$ is learned using sentential and all static KG context \cite{DBLP:conf/emnlp/VashishthJPBT18,bastos2020recon}. However, in \textit{KGPool}, relation $r$ is realized based on the sentential context and dynamically chosen KG context. Hence, we employ context aggregators similar to the baselines (section \ref{sec:baseline}) for jointly learning the enriched KG context in the form of $CG$ and sentential context. 
The final relation is:
%Then the final relation is predicted as: 
\begin{multline}\label{eq_10}
\small
    P(r \mid e_h, e_t, W) =  softmax(\\\mlp(\vec{r^{\prime}}\oplus \vec{e_h}^{\prime} \oplus \vec{e_t}^{\prime} \oplus \vec{\mathcal{W}}^{\prime}\oplus\vec{R}^{\prime}))
\end{multline}
%where $\oplus$ represents concatenation.
%among the respective vectors. 
%\section{KGPool Approach}

\section{Experimental Setup} \label{sec:experiment}
\subsection{Datasets}
%We use two standard datasets (in English): 
We consider two standard datasets (English version): Wikidata dataset \cite{DBLP:conf/emnlp/SorokinG17} and NYT Freebase \cite{DBLP:conf/pkdd/RiedelYM10}. 
%Both datasets were created using distant supervision (associated statastics are in Table \ref{tab:dataset}).
Both datasets were annotated using distant supervision (associated stats are in Table \ref{tab:dataset}). 
%Table \ref{tab:dataset} shows the statistics related to each datasets.
Datasets include 'NA' as one of the target relations. 
%\todo[inline]{what do you mean by "Datasets include 'NA' as one of the target relations.", I donot see any NA in the table}
\begin{table}[h]

   \centering
   \begin{tabular}{p{1.2cm}|p{1.4cm}|p{1.5cm}|p{1.2cm}}
      \toprule
       \textbf{Dataset} & \textbf{\#Train Sentences} & \textbf{\#Test sentences} & \textbf{\#Relations} \\
       \midrule
       Wikidata  & 372,059  & 360,334 & 353  \\
        \midrule
       NYT & 455,771 & 172,448 & 53 \\
        \bottomrule
    \end{tabular}
    \caption{Statistics of the Datasets}
    \label{tab:dataset}
\end{table}
%%%%%%%%%%%%%%%%%%%%%%%%%%%%%%%%%%%%%%%%%\\
%%%%%%%%%%%%%%%%%%%%%%%%%%%%%%%%%%%%%%%%%\\
%%%%%%%%%%%%%%%%%%%%%%%%%%%%%%%%%%%%%%%%%\\
%%%%%%%%%%%%%%%%%%%%%%%%%%%%%%%%%%%%%%%%%\\
%%%%%%%%%%%%%%%%%%%%%%%%%%%%%%%%%%%%%%%%%\\
\subsection{KGPool Configurations} \label{sec:config}
%We configure the \textit{KGPool} model with different context aggregator modules. 
\textit{KGPool} is configured with two context aggregator modules. 
We inherit context aggregators from existing sentential RE baselines. 
%Our idea here is to understand how \textit{KGPool} performs along with already used context aggregators in the literature.
Our experimental aim is to assess as how \textit{KGPool} performs along with the state-of-the-art context aggregators (comparative study). Our two settings are:\\
\textbf{1.} \textbf{$KGPool_{+lstm}$}: \textit{KGPool} is coupled with a context aware LSTM model from \cite{DBLP:conf/emnlp/SorokinG17} as context aggregator.  \\
\textbf{2.} \textbf{$KGPool_{+gnn}$}: this implementation has \textit{KGPool} plugged-in with a variant of GNN module used by \cite{DBLP:conf/acl/ZhuLLFCS19,bastos2020recon}.\\
\subsection{Baseline Models} \label{sec:baseline}
We consider the recent sentential RE approaches for our empirical study:\\
\textbf{RECON} \cite{bastos2020recon}: induces KG context (entity attributes and 1\&2 hop entity triples) along with the sentence in a GNN.\\
%%%%%%%%%%%%%%%%%%%%%%%%%%%%%%%%%%%%%%%%%\\
\textbf{RECON-EAC} \cite{bastos2020recon}: a variant of RECON contains entity attributes as only KG context (same context as \textit{KGPool}). \\
%\textbf{RECON-EAC} \cite{bastos2020recon}: a variant of RECON that excludes context from the triples. Similar to \textit{KGPool}, the entity attributes are the only KG context. \\
%%%%%%%%%%%%%%%%%%%%%%%%%%%%%%%%%%%%%%%%%\\
%\textbf{GP-GNN} \cite{DBLP:conf/acl/ZhuLLFCS19}: performs multi-hop reasoning between the entity nodes for the sentential RE.  \\
\textbf{GP-GNN} \cite{DBLP:conf/acl/ZhuLLFCS19}: performs multi-hop reasoning using a GNN.  \\
%%%%%%%%%%%%%%%%%%%%%%%%%%%%%%%%%%%%%%%%%\\
%%%%%%%%%%%%%%%%%%%%%%%%%%%%%%%%%%%%%%%%%\\
\textbf{Context-LSTM} \cite{DBLP:conf/emnlp/SorokinG17}: uses context from other sentential relations.\\
%%%%%%%%%%%%%%%%%%%%%%%%%%%%%%%%%%%%%%%%%
\textbf{Sorokin-LSTM} \cite{DBLP:conf/emnlp/SorokinG17}: the NYT dataset contains one relation per sentence, but Context-LSTM requires at least two relations in a sentence. Thus, the other setting is an LSTM model without a sentential relation context, is used as a baseline on the NYT dataset.\\
%%%%%%%%%%%%%%%%%%%%%%%%%%%%%%%%%%%%%%%%%
%%%%%%%%%%%%%%%%%%%%%%%%%%%%%%%%%%%%%%%%%
\textbf{Multi-instance RE Approaches}:
%On the NYT dataset, \cite{bastos2020recon} provides a comparison with multi-instance RE approaches. 
Please note, the Wikidata dataset does not have multiple instances for an entity pair. Hence multi-instance baselines do not have values on it. We inherit the recent multi-instance baselines and all empirical values from \cite{bastos2020recon}: (i) \textbf{HRERE} \cite{DBLP:conf/naacl/XuB19} (ii) \textbf{Wu-2019} \cite{DBLP:conf/aaai/WuFZ19}, (iii) \textbf{Yi-Ling-2019} \cite{DBLP:conf/naacl/YeL19}, (iii) \textbf{RESIDE} \cite{DBLP:conf/emnlp/VashishthJPBT18}.
%%%%%%%%%%%%%%%%%%%%%%
%%%%%%%%%%%
%%%%%%%%%%%%%%%%%%%%%%%
%%%%%%%%%%%
%%%%%%%%%%%%%%%%%%%%%%
%%%%%%%%%%%
%%%%%%%%%%%%%%%%%%%%%%%
%%%%%%%%%%%
\begin{figure*}
  \begin{subfigure}[b]{0.41\textwidth}
    \includegraphics[width=\textwidth]{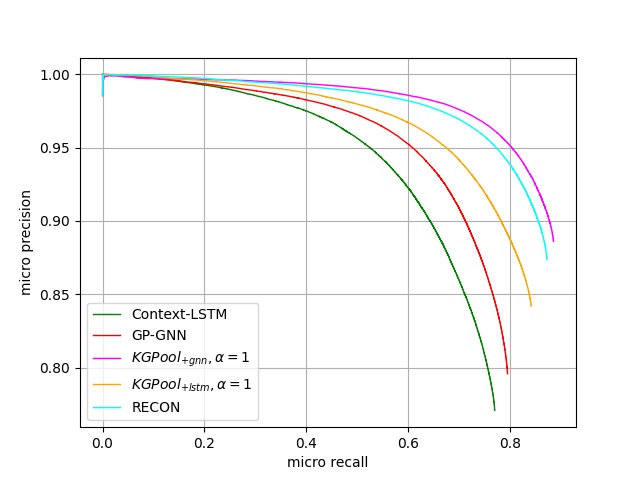}
    \caption{Micro P-R Curve on Wikidata Dataset}
    \label{fig:1}
    % \vspace{-2mm}
  \end{subfigure}
  \begin{subfigure}[b]{0.41\textwidth}
    \includegraphics[width=\textwidth]{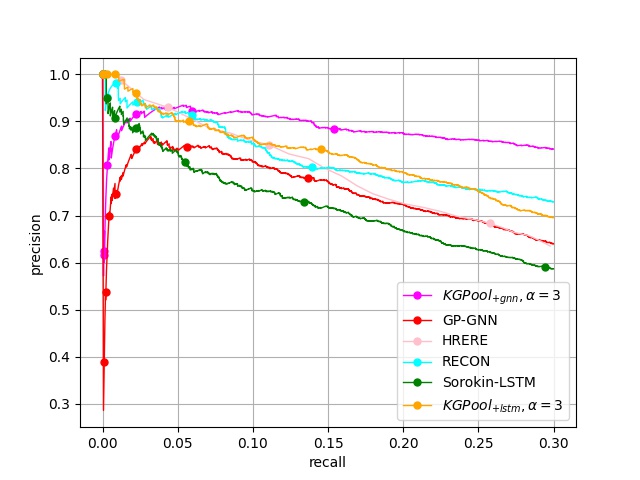}
    \caption{Micro P-R Curve on NYT-Freebase Dataset}
    \label{fig:2}
    % \vspace{-2mm}
  \end{subfigure}
  \caption{\textit{KGPool}'s best configuration (Tables: \ref{tab:results1}, \ref{tab:results2}) perform better than baselines over the entire recall range.}
  \label{fig:prcurve}
      \vspace{-3mm}
\end{figure*}
\subsection{Metrics and Hyper-parameters}
Graph Construction step (section \ref{sec:graph_construction}) use a Bi-LSTM with one hidden layer of size 50 \cite{bastos2020recon}. The word embedding dimension is 50 initialized by Glove embeddings \cite{pennington2014glove}. The context pooling parameters are from \cite{lee2019self}. For model $M$, we used the default parameters provided by the authors \cite{DBLP:conf/acl/ZhuLLFCS19,DBLP:conf/emnlp/SorokinG17}. For brevity, details are in the appendix.\\
%%%%%%%%%%%%%%%%%%%%%%%%%%%%%%%%%%%%%%%%%%%%%%%%%%%%%%%%%%%%%
%%%%%%%%%%%%%%%%%%%%%%%%%%%
\textbf{Metric and Optimization:} Our experiment settings are borrowed from \cite{bastos2020recon}. Hence, on Wikidata dataset, we use (micro) precision (P), recall (R), and F-score (F1). On the NYT Freebase dataset, (micro) P@10 and P@30 is reported. P@K here represents precision at K percent recall. 
%The maximum value can be 100 if the model is very good at detecting all relations correctly for all thresholds. 
We also study the effect of Context Coefficient ($\alpha$) for both \textit{KGPool} configurations (trained end-to-end). We ignore the probability predicted for the NA relation during testing. 
We employ the Adam optimizer \cite{DBLP:journals/corr/KingmaB14} with categorical cross entropy loss where each model is run three times on the whole training set. For the P/R curves (with best $\alpha$ values of \textit{KGPool} variants), the result from the first run of each model is selected. For ablation, we use the McNemar's test for statistical significance to find if the reduction in error in the \textit{KGPool} configurations are significant.
 The differences in the models is statistically significant if the $p-value<0.05$ \cite{dietterich1998approximate}. We release all experiment code and data on a public GitHub\footnote{\url{https://github.com/nadgeri14/KGPool}}.
%For the column "contingency table" (2x2 contingency table), the values of the first row and second column ($RW$) signify the number of instances that model 1 predicted correctly and model 2 incorrectly. The values of the second row and first column gives the number of instances that model 2 predicted correctly and model 1 predicted incorrectly ($WR$). The differences in the models is statistically significant if the $p-value<0.05$ \cite{dietterich1998approximate}. We release all experiment code, datasets and trained models on a public GitHub: \textit{blind-review}.
%The statistic here is 
%\[\frac{(RW-WR)^2}{RW+WR}\]

%% -- Training Details to follow -- %%

%%%%%%%%%%%%%%%%%%%%%%
%%%%%%%%%%%
%%%%%%%%%%%%%%%%%%%%%%%
%%%%%%%%%%%

%%%%%%%%%%%%%%%%%%%%%%
%%%%%%%%%%%
%%%%%%%%%%%%%%%%%%%%%%%
%%%%%%%%%%%
%%%%%%%%%%%%%%%%%%%%%%
%%%%%%%%%%%
%%%%%%%%%%%%%%%%%%%%%%%
%%%%%%%%%%%

%%%%%%%%%%%%%%%%%%%%%%
%%%%%%%%%%%
%%%%%%%%%%%%%%%%%%%%%%%
%%%%%%%%%%%

%%%%%%%%%%%%%%%%%%%%%%
%%%%%%%%%%%
%%%%%%%%%%%%%%%%%%%%%%%
%%%%%%%%%%%

\section{Results}\label{sec:results}
We conduct our experiments and analysis in response to  the question \textbf{RQ}: "What is the efficacy of \textit{KGPool} in dynamically selecting the KG context for the sentential RE task?" As such, we also compare \textit{KGPool} against approaches that do not dynamically treat the context.
%We study the research questions: \textbf{RQ}: "What is the efficacy of \textit{KGPool} in dynamically selecting the KG context for the sentential RE?" As such, we also study if dynamically choosing KG context statistically significant?\todo[inline]{We conduct our experiment and analysis in response to  the question: "What is the efficacy of \textit{KGPool} in dynamically selecting the KG context for the sentential RE task?" As such, we also compare \textit{KGPool} against approaches that do not dynamically treat the context.}\todo[inline]{}

%and \textbf{RQ2}: What is the impact of context pooling on the average degree of entity nodes in $HIG$?
%%%%%%%%%%%%%%%%%%%%%%
%%%%%%%%%%%

\begin{table}[ht!]
    \centering
    % \begin{tabular}{p{1cm}|p{3cm}|p{0.65cm}p{0.65cm}p{0.65cm}|p{0.65cm}p{0.65cm}p{0.65cm}|p{0.65cm}p{0.65cm}p{0.65cm}}
    \begin{tabular}{p{3.5cm}|p{0.7cm}p{0.7cm}p{0.7cm}}
   %  \Cline{1-7}
   \toprule
    %  & & & & & \\
   % &\multicolumn{3}{c|}{}\\
    %   \multicolumn{3}{c}{Micro}\\
     %\cline{2-7}
     \textbf{Model} &P & R & F1  \\
    % \cline{1-7}
    \midrule
    Context-LSTM & 72.09 &72.06 &72.07  \\
    %  \cline{2-7}  
     GP-GNN & 82.30  &82.28  &82.29 \\
      RECON-EAC  & 85.44 & 85.41 & 85.42 \\
      RECON & 87.24 & 87.23 & 87.23 \\
   %  \cline{1-7} 
   \midrule
  $KGPool_{+lstm}$ ($\alpha$=1) & 84.20 & 82.19 & 84.20 \\
   $KGPool_{+lstm}$  ($\alpha$=2) & 84.12 & 84.13 & 84.12 \\
    $KGPool_{+lstm}$  ($\alpha$=3) & 84.00 & 83.97 & 83.98 \\
     $KGPool_{+lstm}$  ($\alpha$=4) & 83.81 & 83.79 & 83.80 \\
    %  \cline{1-2}\cline{3-4}\cline{5-5}\cline{6-7}
    %  & \hspace{15.5mm}+
    \midrule
     $KGPool_{+gnn}$ ($\alpha$=1) & \textbf{88.60} & \textbf{88.59} & \textbf{88.60} \\
       $KGPool_{+gnn}$ ($\alpha$=2) & 88.57 & 88.56 & 88.57 \\
         $KGPool_{+gnn}$ ($\alpha$=3) & 88.54 & 88.55 & 88.54 \\
           $KGPool_{+gnn}$ ($\alpha$=4) & 88.52 & 88.50 & 88.51 \\
    % \Cline{1-7} 
    \bottomrule
 \end{tabular}
\caption{Comparison of \textit{KGPool} configurations with sentential RE models on Wikidata dataset. Best score in bold.}
\label{tab:results1}
    \vspace{-2mm}
\end{table}

\begin{table}[ht!]
    \centering
    % \begin{tabular}{p{1cm}|p{3cm}|p{0.65cm}p{0.65cm}p{0.65cm}|p{0.65cm}p{0.65cm}p{0.65cm}|p{0.65cm}p{0.65cm}p{0.65cm}}
    \begin{tabular}{p{3.5cm}|p{0.7cm}p{0.7cm}}
   %  \Cline{1-7}
   \toprule
    %  & & & & & \\
   % &\multicolumn{3}{c|}{}\\
    %   \multicolumn{3}{c}{Micro}\\
     %\cline{2-7}
     \textbf{Model} &P@10 & P@30  \\
    % \cline{1-7}
    \midrule
    %  \multirow{2}{*}{Baselines} &
    HRERE  & 86.1 & 76.6\\
     Wu-2019  & 81.7 & 61.8\\
     RESIDE & 73.6 &  59.5 \\
   Ye-Ling-2019 & 78.9 & 62.4 \\
    Sorokin-LSTM  & 75.4 & 58.7\\
     GP-GNN   & 81.3 & 63.1\\
        RECON-EAC  &  83.5& 73.4 \\
      RECON & 87.5 &  74.1 \\
   %  \cline{1-7} 
   \midrule
     %\multirow{3}{*}{ReCoN} &
  $KGPool_{+lstm}$ ($\alpha$=1) & 83.7 & 72.7  \\
   $KGPool_{+lstm}$  ($\alpha$=2)  & 83.5 & 71.6 \\
    $KGPool_{+lstm}$  ($\alpha$=3)& 84.1 & 70.6  \\
     $KGPool_{+lstm}$  ($\alpha$=4) & 83.1 & 72.1 \\
    %  \cline{1-2}\cline{3-4}\cline{5-5}\cline{6-7}
    %  & \hspace{15.5mm}+ RECON-EAC-KGGAT 
    \midrule
     $KGPool_{+gnn}$ ($\alpha$=1) & 90.1 &\textbf{86.7}\\
       $KGPool_{+gnn}$ ($\alpha$=2) & 91.0 & 85.0 \\
         $KGPool_{+gnn}$ ($\alpha$=3) & \textbf{92.3} & 85.4\\
           $KGPool_{+gnn}$ ($\alpha$=4) & 90.6 & 84.4 \\
    % \Cline{1-7} 
    \bottomrule
 \end{tabular}
\caption{Comparison of \textit{KGPool} with sentential and multi-instance RE models on NYT Freebase dataset. Best score in bold.}
\label{tab:results2}
    \vspace{-2mm}
\end{table}
\begin{table*}[!htb]
\small
    \centering
    \begin{tabular}{p{4.2cm}|p{1.0cm}|p{1.0cm}|p{1.2cm}|p{2.6cm}|p{1.5cm}|p{1.5cm}}
        \Cline{1-7}
        \textbf{Compared Models} & \multicolumn{2}{|c|}{\textbf{Contingency table}} & \textbf{Statistic} & \textbf{p-value}  & \textbf{Significance} & \textbf{Dataset} \\
        \hline
        % \hline
        $KGPool_{+gnn}$ ($\alpha$=3)  vs  &160916 &4702 &298.18  &$8.18*10^{-67}$ &Statistically & NYT- \\
        %  \Cline{2-3}
        RECON &3169 &3613 &  & &Significant& Freebase \\
          \hline
        %   \hline
         $KGPool_{+gnn}$ ($\alpha$=1) vs &617266 &38652 &1300.08  & $1.08*10^{-284}$&Statistically & Wikidata  \\
        %  \Cline{2-3}$1.08*10^{-284}
          RECON  &29255 &55593 &  & &Significant & \\
          \hline
        %   \hline
        \Cline{1-7}
    \end{tabular}
    \caption{The McNemar's test for statistical significance for KGPool's best configuration Vs previous baseline.}
    \label{mcnemars_test}
    \vspace{-3mm}
\end{table*}
%%%%%%%%%%%%%%%%%%%%%%%%%%%%%%%%%%%%%%%%%%%%%%%%%%%%%%%
%%%%%%%%%%%%%%%%%%%%%%%%%%%%%%%%%%%%%%%%%%%%%%%%%%%%%%%
%%%%%%%%%%%%%%%%%%%%%%%%%%%%%%%%%%%%%%%%%%%%%%%%%%%%%%%
%%%%%%%%%%%%%%%%%%%%%%%%%%%%%%%%%%%%%%%%%%%%%%%%%%%%%%%%%%%%%%%%%%%%%%%%%%%%%%%%%%%%%%%%%%%%%%%%%%%%%%%%%%%%%%%%%%%%%%%%%%%%%%%%%%%%%%%%%%%%%%%%%%%%%%%%%%%%%%%%%%
%%%%%%%%%%%%%%%%%%%%%%%%%%%%%%%%%%%%%%%%%%%%%%%%%%%%%%%%%%%%%%%%%%%%%%%%%%%%%%%%%%%%%%%%%%%%%%%%%%%%%%%%%%%%%%%%%%%%%%%%%%%%%%%%%%%%%%%%%%%%%%%%%%%%%%%%%%%%%%%%%%
%%%%%%%%%%%%%%%%%%%%%%%%%%%%%%%%%%%%%%%%%%%%%%%%%%%%%%%%%%%%%%%%%%%%%%%%%%%%%%%%%%%%%%%%%%%%%%%%%%%%%%%%%%%%%%%%%%%%%%%%%%%%%%%%%%%%%%%%%%%%%%%%%%%%%%%%%%%%%%%%%%
\textbf{Performance on Wikidata Dataset:} 
Table \ref{tab:results1} summarizes the performance of \textit{KGPool} variants against the sentential RE models. Agnostic of the underlying aggregator (LSTM or GNN), \textit{KGPool} effectively captures the KG context complimenting the sentential context. The $KGPool_{+gnn}$ ($\alpha$=1) configuration outperforms other \textit{KGPool} variants along with all sentential RE baselines. We can observe that even when the available context is limited to entity attributes, the $KGPool_{+gnn}$ variant surpasses RECON that also contains context from 1\&2 hop triples besides the entity attributes. RECON-EAC and $KGPool_{+gnn}$ rely on entity attributes as KG context with the same context aggregator. When $KGPool_{+gnn}$ variants choose KG context dynamically, they perform better than RECON-EAC. It is interesting to notice that when an LSTM model is fed with the dynamically chosen context, the performance gain is more than ten absolute points
($KGPool_{+lstm}$ Vs Context-LSTM), even outperforming GP-GNN. \\
%Also, higher value of $\alpha$ signifies more context is fed into the model, resulting in a performance decrease.\\
\textbf{Performance on NYT Freebase Dataset:} Similar to the Wikidata dataset, the $KGPool_{+gnn}$ variants significantly outperform all baselines (cf. Table \ref{tab:results2}). The P@30 is comparatively high for $KGPool_{+gnn}$ against baselines. The behavior could be interpreted as follows: dynamically adding context from the KG for the entity pairs keeps the precision higher over a more extended recall range. For both datasets, \textit{KGPool} configurations ($KGPool_{+gnn}$ and $KGPool_{+lstm}$) have the best-reported performance varying as per the $\alpha$. This validates our choice to introduce a soft constraint in selecting the context nodes (cf., Equation \ref{eq_7}). The P/R curves in Figure \ref{fig:prcurve} show that \textit{KGPool} performs better than baselines over the entire recall range. We conclude that the effective dynamic context selection by \textit{KGPool} has a positive impact on the sentential RE task (which successfully answers our research question).
%\todo[inline]{instead of  (answering \textbf{RQ}), which confirms our hypothesis}
\begin{table}[!htb]
\small
    \centering
    \begin{tabular}{p{3.0 cm}|p{1cm}|p{1cm}|p{1.1cm}}
        \Cline{1-4}
        \textbf{Models} & \textbf{DEG (HIG)}  &  \textbf{DEG (CG)}   & \textbf{Dataset} \\
        \hline
        % \hline
        $KGPool_{+gnn}$ ($\alpha$=1)  & 5.33 & 1.15 & Wikidata \\
         $KGPool_{+gnn}$ ($\alpha$=2)  & 5.33& 1.52 &  \\
          $KGPool_{+gnn}$ ($\alpha$=3)  & 5.33 & 2.87 & \\
           $KGPool_{+gnn}$ ($\alpha$=4)  & 5.33 & 4.71 &  \\
        %  \Cline{2-3}
          \hline
        %   \hline
       $KGPool_{+gnn}$ ($\alpha$=1)  & 6.34 & 1.67 & NYT \\
         $KGPool_{+gnn}$ ($\alpha$=2)  &6.34& 1.91 &  Freebase\\
          $KGPool_{+gnn}$ ($\alpha$=3)  & 6.34 & 2.73 & \\
           $KGPool_{+gnn}$ ($\alpha$=4)  & 6.34 & 5.16 &  \\
        %  \Cline{2-3}
          \hline
        %   \hline
        \Cline{1-4}
    \end{tabular}
    \caption{Effect of Context Pooling. ‘DEG' denotes average degree of an entity node ($e_i$). ‘DEG' of entity nodes in $CG$ is drastically reduced wrt the $HIG$.}
    \label{degree_of_nodes}
    \vspace{-3mm}
\end{table}
%%%%%%%%%%%%%%%%%%%%%%%%%%%%%%%%%%%%%
%%%%%%%%%%%%%%%%%%%%%%%%%%%%%%%%%%%%%
%%%%%%%%%%%%%%%%%%%%%%%%%%%%%%%%%%%%%%%%%%%%%%%%%%%%%%%%%%%%%%%%%%%%%%%%%%%%%%%%%%%%%%%%%%%%%%%%%%%%%%%%%%%%%%%
%%%%%%%%%%%%%%%%%%%%%%%%%%%%%%%%%%%%%%%%%%%%%%%%%%%%%%%%%%%%%%%%%%%%%%%%%%%%%%%%%%%%%%%%%%%%%%%%%%%%%%%%%%%%%%%%%%%%%%%%%%%%%%%%%%%%%%%%%%%%%%%%%%%%%%%%%%%%%%%%%%%%%%%%%%%%%%%%%%%%%%%%%%%%%%%%%%%%%%%%%%%%%%%%%%%%%%%%%%%%%%%%%%%%%%%%%%%%%%%%%%%%%%%%%%%%%%%%%%%%%%%%%%%%%%%%%%%%%%%%%%%%%%%%%%%%%%%%%%%%%%%%%%%%%%%%%%%%%%%%%%%%%%%%%%%%%%%%%%%%%%%%%%%%%%%%%%%%%%%%%%%%%%%%%%%%%%%%%%%%%%%%%%%%%%%%%%%%%%%%%%%%%%%%%%%%%%%%%%%%%%%%%%%%%%%%%%%%%%%%%%%%%%%%%%%%%%%%%%%%%%%%%%%%%%%%%%%%%%%%%%%%%%%%%%%%%%%%%%%%%%%%%%%%%%%%%%%%%%%%%%%%%%%%%%%%%%%%%%%%%%%%%%%%%%%%%%%%%%%%%%%%%%%%%%%%%%%%%%%%%%%%%%%%%%%%%%%%%%%%%%%%%%%%%%%%%%%%%%%%%%%%%%%%%
\subsection{Ablation Studies}
We conducted two ablation studies to understand the behavior of \textit{KGPool} configurations:\\
\textbf{Significance of Dynamic Context Selection:} we perform McNemar's test for the best \textit{KGPool} configuration against the previous sentential state-of-the-art (i.e. RECON). The results in Table \ref{mcnemars_test} are statistically significant on both datasets, illustrating \textit{KGPool}'s robustness. Although $KGPool_{+gnn}$ variants achieve statistically significant results against RECON, there exist several sentences for which our approach is unable to select supplementary KG context (($RW$) values in the contingency table). It requires further investigation, and we plan it for our future work.\\
\textbf{Effect on the Degree of Nodes for Entities:} for studying the effect of context pooling (Section \ref{sec:context_pooling}), we also conducted a study to understand the impact of \textit{KGPool} on the reduction of the average degree of entity nodes ($e_i$) in the $HIG$. Table \ref{degree_of_nodes} summarizes the effect of Context Coefficient on the average degree of entity nodes. Irrespective of $\alpha$, \textit{KGPool} notably reduces the degree of $e_i$ by removing less relevant nodes. \\
\textbf{Architectural Choice Experiment:}
In \textit{KGPool}, we chose to introduce pooling in the last layer of a three-layered architecture (three blocks). To support our choice, we performed several additional experiments by introducing pooling in various layers. We employ the Wikidata dataset for our experiments. We use best configuration of our model ( $KGPool_{+gnn}$ ($\alpha$=1)) and created several variants of it. For instance,  $KGPool_{+gnn}$ ($P$=all) comprises the configuration where we introduce pooling in all three GCN blocks. The configuration $KGPool_{+gnn}$ ($\P$=2\&3) has no pooling in the first layer but has a pooling layer in the remaining two GCN blocks.  $KGPool_{+gnn}$ is the best configuration of \textit{KGPool} where pooling is just in the final layer. In Table \ref{tab:pooling}, we observe that $KGPool_{+gnn}$ with pooling only in the last GCN block has the superior performance compared to other two variants.
Here, the first two layers are used to learn the node features, which are then employed with self-attention for node selection. Our experiments justify the architectural choice decision. However, with a newer graph pooling technique, such decisions will solely depend on the performance of the approach, and we can not generalize the results of these experiments. 
\begin{table}[!htb]
    \centering
   \begin{tabular}{p{3.5cm}|p{0.8cm}}
       % \Cline{1-4}
      \hline
      %  & &  & \\
       % & & & \\
      \textbf{Model}  & \textbf{F1} \\
        \Cline{1-2}
      %\midrule
       % GP-GNN &  &  &  \\
     %\multirow{3}{*}{ReCoN} & GP-GNN + KBGAT  
  $KGPool_{+gnn}$ ($\P$=all) & 84.19 \\
    %  \cline{1-2}\cline{3-4}\cline{5-5}\cline{6-7}
    %  & \hspace{15.5mm}+ RECON-EAC-KGGAT & 
     $KGPool_{+gnn}$  ($\P$=2\&3)  & 86.87 \\
          $KGPool_{+gnn}$ ($\P$=3) (best) & \textbf{88.60} \\
       \hline
    \end{tabular}
    \caption{When we introduce pooling in all three layers or in two layers, the performance of \textit{KGPool}'s variants drop. Hence, it justify our choice to add pooling only in the third layer that gives the best performance (values in bold). We use best configuration of our model ($KGPool_{+gnn}$ ($\alpha$=1)).}
    \label{tab:pooling}
        \vspace{-2mm}
\end{table} 
\\
\textbf{Case-Studies:}
To understand the \textit{KGPool}'s performance gain, we report a few top relations in Table \ref{best}. It can be observed from this table that in a few cases, with lesser context, \textit{KGPool} can perform significantly better. In the next case study, to understand the \textit{KGPool}'s performance while adding additional context (more noise), we induce extra context in the form of 1\&2-hop triples along with entity attributes. For the same, we considered \textit{KGPool}'s best configurations on the Wikidata dataset. The configurations $KGPool_{+gnn}$ ($+T$) and $KGPool_{+lstm}$ ($+T$) represent \textit{KGPool} fed with additional triple context. For both configurations agnostic of underlying aggregator, we observe a slight increase in performance (Table \ref{tab:triple}). There are several triples which are the irrelevant source of information not needed for a given sentence. \textit{KGPool} can remove that information and does not suffer the performance drop due to added noise in the context. Details on error analysis, performance for worst performing individual relations, and on a human-annotated dataset are provided in the appendix.
%To further improve the triple-context impact, including edge features is a viable future step.
\\
\begin{table}[!htbp]
\small
    \centering
    \begin{tabular}{p{2.5 cm}|p{1.2cm}|p{1.2cm}|p{1.3cm}}
        \Cline{1-4}
        \textbf{Relation} & \textbf{\textit{KGPool}}  &  \textbf{RECON}&  \textbf{GP-GNN}  \\
        \hline
        % \hline
          vocal specialization & \textbf{1.00} & 0.00& 0.00\\
          list of works &\textbf{1.00} & 0.00& 0.00\\
          track gauge & \textbf{1.00} & 0.92 & 0.00\\
         position played & \textbf{0.99} & \textbf{0.99}& 0.92 \\
          sport & \textbf{0.99} & \textbf{0.99} & 0.97\\
          record label &\textbf{0.95} & 0.90& 0.64\\
          list of episodes & \textbf{0.95} & 0.00& 0.49\\
        wing configuration & \textbf{0.94} & 0.57& 0.00 \\
       numeric value & \textbf{0.93} & 0.27 & 0.46\\
        vessel class & \textbf{0.87} & 0.00& 0.00 \\
        
        %   \hlin
        %  \Cline{2-3}
          \hline
        %   \hline
        \Cline{1-4}
    \end{tabular}
   % \caption{Micro F-score of randomly selected Top performing Relations for $KGPool_{gnn}$ ($\alpha$=1) (from Wikidata dataset). We observe that when context is dynamically chosen, many relations have significant performance jump compared to RECON.}
   \caption{Micro F-score of Top performing Relations for $KGPool_{gnn}$ ($\alpha$=1) (on Wikidata dataset). Dynamically chosen context significantly improves performance for many relations.}
    \label{best}
    \vspace{-3mm}
\end{table}
%The case study allows us to dig deeper into the jump we observed in the main paper's empirical studies.

\begin{table}[!htb]
    \centering
   \begin{tabular}{p{3.5cm}|p{0.8cm}}
       % \Cline{1-4}
      \hline
      %  & &  & \\
       % & & & \\
      \textbf{Model}  & \textbf{F1} \\
        \Cline{1-2}
      %\midrule
       % GP-GNN &  &  &  \\
     %\multirow{3}{*}{ReCoN} & GP-GNN + KBGAT  & 
  $KGPool_{+gnn}$ ($+T$) & \textbf{88.85} \\
   $KGPool_{+lstm}$ ($+T$) & 84.42 \\
   \hline
    %  \cline{1-2}\cline{3-4}\cline{5-5}\cline{6-7}
    %  & \hspace{15.5mm}+ RECON-EAC-KGGAT & 
          $KGPool_{+gnn}$ & 88.60 \\
            $KGPool_{+lstm}$ & 84.12 \\
       \hline
    \end{tabular}
    \caption{To scale the sources of the contexts, we induce additional triple context in the \textit{KGPool} shown as ($+T$) configurations. We use best configurations of our model ($KGPool_{+gnn}$ ($\alpha$=1) and  $KGPool_{+lstm}$ ($\alpha$=1)). We observe a slight jump in the performance, however, \textit{KGPool} is still able to pool irrelevant context.}
    \label{tab:triple}
        \vspace{-2mm}
\end{table}

%%%%%%%%%%%%%%%%%%%%%%%
%%%%%%%%%%%

\section{Discussion and Conclusion} \label{sec:conclusion}
%Although KGs are used as background knowledge in the RE task (cf. section \ref{sec:related}), yetlittle is known about which context is relevant for the input sentence.
Although KGs are often employed for providing background context in the RE tasks (cf. Section \ref{sec:related}), yet there is limited research about defining relevant context.
In this work, we proposed \textit{KGPool} and provide a set of experiments proving: 1) Given the limited context that is in individual sentences, dynamically bringing context from KG significantly improves the RE performance.
%when the context is limited to a sentence, mapping the context selection task as Graph Pooling problem helps to dynamically select the KG context for significantly improving the RE performance.
%\todo[inline]{1) Given the short context that happens in individual sentences, dynamically bringing context from KG significantly improves the RE performance.}
2) We introduced Context Coefficient ($\alpha$), which acts as a soft constraint in determining the relevant entity context nodes.
%2) Context Coefficient ($\alpha$) acts as a soft constraint in determining the relevant entity context nodes. \todo[inline]{2) We introduced Context Coefficient ($\alpha$), which acts as a soft constraint in determining the relevant entity context nodes.}
3) Our approach \textit{KGPool} is invariant of the context aggregator and enables us to learn effective knowledge representation of the required KG context for a given sentential context.
%3) \textit{KGPool} is invariant of the context aggregator and able to learn effective knowledge representation of the required KG context for a given sentential context. \todo[inline]{3) Our approach \textit{KGPool} is invariant of the context aggregator and enables us to learn effective knowledge representation of the required KG context for a given sentential context. }
Our evaluation concerns several key questions:
\begin{itemize}
\vspace{-2mm}
  \item \textbf{Data quality impact on an effective knowledge representation:} in spite \textit{KGPool}'s significant performance, there exist several sentences for which our model finds a limitation compared to the baseline (cf. Table \ref{mcnemars_test}). One potential interpretation could be about the noise injected due to the data quality of the KG context \cite{weichselbraun2018mining}. Hence, how does the quality of contextual data impact the performance of context selection approaches is an open direction.
  \item \textbf{Impact of additional sources of KG context:} In ablation, we provide a study by adding 1 \& 2-hop triples in addition to entity attributes. There is no significant increase in the performance, although \textit{KGPool} is able to remove irrelevant context for a given sentence. Furthermore, we did not consider edge features in $HIG$ although \textit{KGPool} can be extended to support edge features using techniques such as \cite{DBLP:conf/cvpr/SimonovskyK17}. Additional experiments are needed to verify that our empirical observations hold in this setting, and we leave it for future work.
\end{itemize}

Overall, \textit{KGPool} provides an effective knowledge representation for set-ups where sentence context is sparse. It is interesting to observe that effective knowledge representation learned using \textit{KGPool} paired with an LSTM model outperforms GP-GNN \cite{DBLP:conf/acl/ZhuLLFCS19}, and nearly all multi-instance baselines. Our conclusive results open a new research direction: \textit{is it possible to apply effective context selection techniques coupled with deep learning models to other downstream NLP tasks?} For example, our results can encourage researchers to extend \textit{KGPool} or develop novel context selection methods for the tasks where KGs have been extensively used as additional background knowledge, such as in entity linking \cite{mulang2020evaluating,mulang2020encoding}, KG completion \cite{wang2020entity,shi2017knowledge}, and recommendation system \cite{yang2020contextualized}. 
%Another open direction is to scale \textit{KGPool} in an industrial setting and further improve the node selection techniques. 

\section{Ethics/ Impact Statement:}
In this work, we present significant progress in solving sentential RE task. 
Harvesting knowledge is an essential goal that human beings seek along with the advancement of technology. This research and many RE approaches rely on additional signals from the public KGs to design systems that extract structured knowledge from unstructured contents. When it comes to who may be disadvantaged from this research, we do not think it is applicable
since our study of addressing the KG context capabilities is still at an early stage.
Having said so,
we are fully supporting the development of ethical and responsible AI. The potential bias in the standard public datasets
that may lead to wrong knowledge needs to be cleaned or corrected with validation mechanisms. 
\bibliographystyle{acl_natbib}

\bibliography{emnlp2018} 
\appendix

% =====================================================
% =====================================================
% =====================================================
% =====================================================
\section{Appendix}
Due to page limit, we could not put several empirical results in the main paper. This section describes the remaining empirical studies. 
\subsection{Error Analysis}
To understand the failure cases of \textit{KGPool}, we conducted exhaustive error analysis. We calculated (micro) F1-Score of each relation in Sorokin dataset \cite{DBLP:conf/emnlp/SorokinG17}. Table \ref{worst} illustrates performance of ten relations on which \textit{KGPool} performs the worst (ascending order of Micro-F1 score). To put the study in the right perspective, we also report all sentential RE baselines' performance on these relations. While analyzing the errors, we observe three patterns. First, all models fail in the relations for which number of instances are sparse. For example, the relation \textit{mother} has only 190 instances (occurances) and the relation \textit{killed by} has 48 instances. The scarcity in training data has made all models to fail on certain relations. 
Secondly, our model fails in very closed relations. For example, instead of predicting the relation \textit{drafted by}\footnote{\url{https://www.wikidata.org/wiki/Property:P647}}, our model predicts \textit{member of sport team}\footnote{ \url{https://www.wikidata.org/wiki/Property:P54}}. Similarly, in case of \textit{unmarried partner}, our model predicts \textit{spouse}. We believe that introducing logical reasoning in the model can help these borderline cases. The third observed pattern for errors is the quality of context. It is worthwhile to mention that in GP-GNN and Context-LSTM, there is only a sentential context. RECON and \textit{KGPool} use KG context. Still, performance is limited for many relations such as \textit{use} and \textit{different from} as reported in the table \ref{worst}. The lack of quality context in the KG possibly a reason for limited performance for KG-context-induced models in erroneous cases. Detailed exploration is needed to understand the impact of data quality on \textit{KGPool} performance, and we leave it for the future work. 
%We release performance of \textit{KGPool} including the baseline models for individual relations of both datasets on our public Github. 

\subsection{Effect of Context Pooling}
\begin{table}[!htb]
\small
    \centering
    \begin{tabular}{p{3.0 cm}|p{1cm}|p{1cm}|p{1.1cm}}
        \Cline{1-4}
        \textbf{Models} & \textbf{DEG (HIG)}  &  \textbf{DEG (CG)}   & \textbf{Dataset} \\
        \hline
        % \hline
        $KGPool_{+lstm}$ ($\alpha$=1)  & 5.33 & 1.06 & Wikidata \\
         $KGPool_{+lstm}$ ($\alpha$=2)  & 5.33& 2.12 &  \\
          $KGPool_{+lstm}$ ($\alpha$=3)  & 5.33 & 4.32 & \\
           $KGPool_{+lstm}$ ($\alpha$=4)  & 5.33 & 4.81 &  \\
        %  \Cline{2-3}
         \hline
        %   \hline
       $KGPool_{+lstm}$ ($\alpha$=1)  & 6.34 & 1.23 & NYT \\
         $KGPool_{+lstm}$ ($\alpha$=2)  &6.34& 1.74 &  Freebase\\
          $KGPool_{+lstm}$ ($\alpha$=3)  & 6.34 & 3.05& \\
           $KGPool_{+lstm}$ ($\alpha$=4)  & 6.34 & 6.30 &  \\
        %  \Cline{2-3}
          \hline
        %   \hline
        \Cline{1-4}
    \end{tabular}
    \caption{Effect of Context Pooling. ‘DEG' denotes average degree of an entity node ($e_i$). We observe a reduction in the degree of entity nodes in $CG$ compared to the $HIG$.}
    \label{degree_of_nodes_lstm}
    \vspace{-3mm}
\end{table}

\begin{table*}[!htbp]
\small
    \centering
    \begin{tabular}{p{3.0 cm}|p{1.5cm}|p{1.5cm}|p{1.5cm}|p{3.1cm}}
        \Cline{1-5}
        \textbf{Relation} & \textbf{\textit{KGPool}}  &  \textbf{RECON}   & \textbf{GP-GNN} & \textbf{Context-LSTM} \\
        \hline
        % \hline
        has quality   & 0.00 & 0.00 & 0.00 & 0.00\\
        enclave within   & 0.00 & 0.00 & 0.00 & 0.00 \\
          drafted by  & 0.01 & 0.08 & 0.00 & 0.02 \\
           different from  & 0.01 & 0.00& 0.0 & 0.00\\
           mother  & 0.03 & 0.05 & 0.02 & 0.00\\
          unmarried partner & 0.04 & 0.01 & 0.00 & 0.00\\
           killed by & 0.04 & 0.01 & 0.00& 0.04\\
           use & 0.09 & 0.00 & 0.00 & 0.00\\
           lyrics by & 0.10 & 0.13 & 0.00 & 0.00\\
          relative & 0.12 & 0.10 & 0.00 & 0.00\\
         % \hline
        %   \hlin
        %  \Cline{2-3}
          \hline
        %   \hline
        \Cline{1-5}
    \end{tabular}
    \caption{Micro F-score of 10-worst performing Relations for $KGPool_{gnn}$ ($\alpha$=1) on Wikidata dataset. We also provide corresponding values of other sentential RE baselines. The main reason for limited performance across all models is the scarcity of training data for these relation types.}
    \label{worst}
    \vspace{-3mm}
\end{table*}

In the main paper, we presented the effect of context pooling on \textit{KGPool}'s best configuration ($KGPool_{+gnn}$). Table \ref{degree_of_nodes_lstm} describes the reduction in the average degree of nodes for $KGPool_{+lstm}$ configuration for various context coefficient ($\alpha$). On both datasets, there is a significant reduction in the degree of nodes. On Wikidata dataset \cite{DBLP:conf/emnlp/SorokinG17}, $KGPool_{+lstm}$ with 
($\alpha$=1) reports the highest value among its other configurations. For the same, the average degree of nodes is reduced from 5.33 to 1.06. Please note, the degree of nodes in $HIG$ remains the same. However, for $CG$, the degree of nodes differs based on the context aggregator. We train the model end to end, and due to back-propagation, context weights adjust as per the context aggregator. 
\subsection{Results on a Human Annotated Dataset}
The employed datasets Wikidata \cite{DBLP:conf/emnlp/SorokinG17} and NYT Freebase  \cite{DBLP:conf/pkdd/RiedelYM10} are created using distant supervision techniques. Considering distant supervision techniques inherit a noise, to provide a comprehensive ablation study, \cite{DBLP:conf/acl/ZhuLLFCS19} provided a human evaluation setting. Following the same setting, RECON provided human-annotated data from Wikidata dataset \cite{DBLP:conf/emnlp/SorokinG17}. This is to verify that the distantly supervised dataset is correct for every pair of entities. Sentences accepted by all annotators are part of the human-annotated dataset. There are 500 sentences and 1846 triples in the test set. Table \ref{tab:human} reports $KGPool$'s performance against the sentential baselines. $KGPool_{+gnn}$ continues to outperform the baselines, maintaining similar behavior as seen on test sets of original datasets. The results further re-assure the robustness of our proposed approach. 
\begin{table}[!htb]
    \centering
   \begin{tabular}{p{3.5cm}|p{0.8cm}|p{0.8cm}|p{0.8cm}}
       % \Cline{1-4}
      \hline
      %  & &  & \\
        & & & \\
      \textbf{Model} & \textbf{P} & \textbf{R} & \textbf{F1} \\
        \Cline{1-4}
      %\midrule
       % GP-GNN &  &  &  \\
         Context Aware LSTM  & 77.77 & 78.69 & 78.23 \\
       GP-GNN & 81.99 & 82.31 & 82.15 \\
        RECON-EAC & 86.10 & 86.58 & 86.33 \\
           RECON& 87.34 & 87.55 & 87.44 \\
       \hline
     %\multirow{3}{*}{ReCoN} & GP-GNN + KBGAT  &
  $KGPool_{+lstm}$ ($\alpha$=1) & 86.34 & 86.07 & 86.20\\
    %  \cline{1-2}\cline{3-4}\cline{5-5}\cline{6-7}
    %  & \hspace{15.5mm}+ RECON-EAC-KGGAT & 
     $KGPool_{+gnn}$ ($\alpha$=1) & \textbf{89.36} & \textbf{89.31} & \textbf{89.33} \\
       \hline
    \end{tabular}
    \caption{Sentential RE performance on Human Annotation Dataset. \textit{KGPool} again outperforms the baselines. We report Micro P,R, and F1 values. (Best score in bold)}
    \label{tab:human}
        \vspace{-2mm}
\end{table} 
\begin{table}[!htb]
\small
    \centering
    \begin{tabular}{p{3cm}p{1cm}}
        \Cline{1-2}
         \textbf{Hyperparameters} &\textbf{Value} \\
        \hline
        learning rate &0.001\\
        \hline
        batch size &50\\
        \hline
        hidden state size &128\\
        \hline
        context  coefficient ($\alpha$) &1,2,3,4\\
        \hline
        $\#$ of propagation layers &3\\
        \hline
        \Cline{1-2}
    \end{tabular}
    \caption{Hyper-parameters for Context Pooling module}
    \label{tab:tab_hyp3}
    \vspace{-3mm}
\end{table}

\begin{table}[!htb]
\small
    \centering
    \begin{tabular}{p{3cm}p{1cm}}
        \Cline{1-2}
         \textbf{Hyperparameters} &\textbf{Value} \\
        \hline
        learning rate &0.001\\
        \hline
        batch size &50\\
        \hline
        dropout ratio &0.5\\
        \hline
        hidden state size &256\\
        \hline
        non-linear activation &relu\\
        \hline
        $\#$ of propagation layers &3\\
        \hline
        entity embedding size &8\\
        \hline
        adjacent matrices &untied\\
        \hline
        optimizer &adam\\
        \hline
        $\beta_1$ &0.9\\
        \hline
        $\beta_2$ &0.999\\
        \hline
        $\varepsilon$ &1e-08\\
        \hline
        pretrained embeddings &glove\\
        \hline
        word embedding dim &50\\ 
        \hline
        \Cline{1-2}
    \end{tabular}
    \caption{Hyper-parameters for GNN-Aggregator module}
    \label{tab:tab_hyp1}
    \vspace{-3mm}
\end{table}

\begin{table}[!htb]
\small
    \centering
    \begin{tabular}{p{3cm}p{1cm}}
        \Cline{1-2}
         \textbf{Hyperparameters} &\textbf{Value} \\
        \hline
        learning rate &0.001\\
        \hline
        batch size &50\\
        \hline
        dropout ratio &0.5\\
        \hline
        hidden state size &256\\
        \hline
        non-linear activation &relu\\
        \hline
        $\#$ of layers &1\\
        \hline
        optimizer &adam\\
        \hline
        pretrained embeddings &glove\\
        \hline
        word embedding dim &50\\ 
        \hline
        \Cline{1-2}
    \end{tabular}
    \caption{Hyper-parameters for ContextAware-Aggregator module}
    \label{tab:tab_hyp1}
   % \vspace{-3mm}
\end{table}
\begin{table}[!htb]
\small
    \centering
    \begin{tabular}{p{3cm}p{1.5cm}}
        \Cline{1-2}
         \textbf{Hyperparameters} &\textbf{Value} \\
        \hline
        learning rate &0.001\\
        \hline
        batch size &50\\
        \hline
        initial embedding size &50\\
        \hline
        final embedding size &50\\
        \hline
        pretrained embeddings &glove\\
        \hline
        $\#$ of layers &1\\
        \Cline{1-2}
        \hline
    \end{tabular}
    \caption{Hyper-parameters for Graph Construction module}
    \label{tab:tab_hyp2}
  %  \vspace{-3mm}
\end{table}

% ===========================================================================================================================================

% ==========================================================================================================================================

     %\end{tabular}
     %\caption{Micro F-score of randomly selected Top performing Relations for $KGPool_{gnn}$ ($\alpha$=1) (from Wikidata dataset). We observe that when context is ynamically chosen, many relations have significant performance jump compared to RECON.}
  %  \label{best}
 %   \vspace{-3mm}
%\end{table}

\subsection{Datasets and Hyper-parameters}
We augmented two datasets Wikidata dataset and Riedel Freebase dataset with our proposed KG context. The Wikidata dataset has 353 unique relations, 372,059 sentences in training, 123824 sentences in validation and  360,334 for testing. The number of sentences in the training and test set are 455,771 and  172,448  respectively in the Riedel dataset. No explicit validation set has been provided for Riedel dataset. For augmenting entity attribute context, we relied on public dumps of Wikidata and Freebase. From these dumps, we automatically extracted entities and its properties: labels, aliases, instance of, descriptions. For Wikidata, we used public API\footnote{\url{https://query.wikidata.org/}} using a SPARQL query and for Freebase, we took original depreciated dump\footnote{\url{https://developers.google.com/freebase}}.

We use the nltk english tokenizer for splitting the sentence into its corresponding tokens in the Riedel dataset. We do not do any further data preprocessing. We used 1 GPU NVIDIA TITAN X Pascal with 12GB of GPU storage to run our experiments. We train the models upto a maximum of 14 epochs and select the best performing model based on the micro F1 scores of the validation set. The tables \ref{tab:tab_hyp1}, \ref{tab:tab_hyp2} and \ref{tab:tab_hyp3} detail the hyper-parameter settings used in our experiments. We do not do any further hyper-parameter tuning. 
%Note that these are the F1 scores ignoring the NA/P0 labels for selection of the best epoch to use and will differ from the exact test metrics reported for comparison with other methods. 
%The detailed model statistics are in \ref{tab:tab_misc_stats2}.
\end{document}